\documentclass[conference]{IEEEtran}
\IEEEoverridecommandlockouts
\usepackage{float} 
\usepackage{booktabs}   
\usepackage{multirow}
\usepackage{makecell}
\usepackage{tabularx}
\usepackage{array}
\usepackage{colortbl}   
\usepackage{subcaption}
\usepackage{amsmath}
\usepackage{amsfonts}
\usepackage{diagbox}
\usepackage{graphicx}
\usepackage{tikz}
\usepackage{overpic}
\usepackage{pgfplots}
\pgfplotsset{compat=1.17}
\usepackage{soul}
\usepackage{comment}
\usepackage{nicematrix,tikz}
\usepgfplotslibrary{colormaps,patchplots}
\usepackage{contour}
\contourlength{0.6pt}
\definecolor{lightblue}{RGB}{252,254,255}
\definecolor{midblue3}{RGB}{236,246,254}
\definecolor{midblue2}{RGB}{210,232,250}
\definecolor{midblue1}{RGB}{170,210,242}
\definecolor{darkblue}{RGB}{120,170,225}
\usepackage{subfig}     

\usepackage{xcolor}
\usepackage{soul}       
\usepackage{gensymb}    

\usepackage{xspace}
\usepackage{verbatim}
\usepackage{comment}
\usepackage{enumitem}
\usepackage{accents}
\usepackage{url}
\usepackage{pifont}
\usepackage[normalem]{ulem} 

\usepackage{graphicx}   
\usepackage{booktabs}
\usepackage[table]{xcolor}
\usepackage{tabularx}
\usepackage{array}
\newcolumntype{C}{>{\centering\arraybackslash}X} 
\definecolor{darkblue}{RGB}{0,51,102}
\definecolor{lightblue}{RGB}{230,240,255}

\usepackage[most]{tcolorbox}
\usepackage{soul}    
\usepackage{xcolor}  

\begin{document}

\title{Auditing Training Data in Generative Music Models\\ via Black-Box Membership Inference}
\author{
\IEEEauthorblockN{
Yi Chen Liu\IEEEauthorrefmark{1},
Jiawei Yu\IEEEauthorrefmark{1},
Kexin Cao\IEEEauthorrefmark{2},
Syed Irfan Ali Meerza\IEEEauthorrefmark{3},
Trishika Movva\IEEEauthorrefmark{1},
Jian Liu\IEEEauthorrefmark{1}
}
\IEEEauthorblockA{\IEEEauthorrefmark{1}University of Georgia, Athens, GA, USA}
\IEEEauthorblockA{\IEEEauthorrefmark{2}Independent Researcher}
\IEEEauthorblockA{\IEEEauthorrefmark{3}University of Tennessee, Knoxville, TN, USA}
\IEEEauthorblockA{Emails: yl51148@uga.edu; Jiawei.Yu@uga.edu; c72919571@gmail.com; \\
smeerza@vols.utk.edu; Trishika.Movva@uga.edu; jianliu@uga.edu}
}
\maketitle

\begin{abstract}
Recent advances in text-to-music generation enable high-fidelity synthesis of structured musical audio, raising growing concerns about data provenance, consent, and training transparency. These models are typically trained on large-scale corpora with little disclosure, leaving no practical mechanism to verify whether a particular audio sample was included in training.
In this paper, we investigate black-box membership inference for generative music models, aiming to determine whether a candidate music sample was used during training, given only query access to the deployed system.
Our key insight is that training membership induces systematically stronger semantic and structural alignment between a candidate sample and the model’s generation conditioned on its caption. We query the target model with the associated caption and measure the relationship between the candidate audio and the generated output in a learned feature space.
To capture features that separate members from non-members, we construct paired examples consisting of each track and its caption-conditioned generation from shadow models, and train a music auditor to classify membership. The auditor captures alignment patterns characteristic of training membership and generalizes to unseen target models in a fully black-box setting without access to model parameters or training metadata.
Across multiple state-of-the-art music generators, our method achieves up to 98.6\% accuracy, with false-positive and false-negative rates as low as 1.9\% and 1.0\%, demonstrating that reliable training-data auditing is feasible in realistic deployment scenarios.

\end{abstract}

\renewcommand{\IEEEkeywordsname}{Keywords}

\vspace{-2mm}
\section{Introduction}
\vspace{-2mm}
Music has traditionally been regarded as a uniquely human form of creativity grounded in cultural context and performance. Recent advances in generative modeling, however, enable high-fidelity music synthesis from minimal input, producing coherent melodies, harmonies, instrumentation, and vocals. Systems such as MusicGen~\cite{copet2023simple}, MusicLM~\cite{agostinelli2023musiclm}, and Lyria RealTime~\cite{lyriateam2025livemusicmodels}, along with commercial platforms like Stable Audio~\cite{evans2025stable}, Suno~\cite{sunoai2023}, and Udio~\cite{udio2024}, demonstrate the growing capability of AI to generate complete, stylistically consistent musical tracks.

Despite their impressive capabilities, the training practices of generative music models remain largely opaque. These models are typically trained on massive audio corpora aggregated from online sources, professional recordings, and user-generated content, often without transparent documentation of data provenance or consent~\cite{hardinges2024we}.
As a result, artists, copyright holders, and platform regulators face a fundamental challenge: \textit{how can one verify whether a specific piece of music has been included in the training of a deployed generative model?} Addressing this question is essential for accountability, consent verification, and responsible deployment of generative music systems.

Membership inference seeks to determine whether a given data sample was used during model training and has been extensively studied in domains such as image classification~\cite{hu2022membership}, language modeling~\cite{duan2024membership}, and text-to-image generation~\cite{wu2022membership}. In contrast, membership inference for generative music models has only emerged very recently. Some works rely on strong assumptions that are impractical in deployed systems, such as teacher-forcing likelihood access for symbolic generators~\cite{liu2026ts} or developer-side white-box access to diffusion states~\cite{liu2026mi}, which require internal model access and are not applicable to real-world black-box deployments. Other recent studies explore black-box auditing for generative audio systems. Chow \textit{et al.}~\cite{chow2025assessing} evaluate existing membership inference attacks on MuseGAN using shadow training and scalar statistics such as discriminator scores and reconstruction errors. Proboszcz \textit{et al.}~\cite{proboszcz2025membership} train logistic regression models on extracted feature vectors using predefined train–test splits of public datasets. However, these approaches primarily rely on weak scalar metrics or fixed dataset partitions and do not explicitly model semantic alignment between original audio and generated outputs. As a result, they struggle to provide reliable per-sample auditing in realistic black-box settings.

Due to memorization phenomena observed in generative models~\cite{somepalli2023understanding, carlini2023extracting}, samples seen during training can induce generation behaviors that are more semantically aligned with the original data. In the context of music generation, if an audio sample was part of the training set, the model's output conditioned on its caption may exhibit stronger semantic consistency with that original audio compared to unseen samples. Motivated by this insight, we propose an embedding-based black-box auditing framework that measures the alignment between a candidate audio sample and the music generated by the target model using its associated (or automatically extracted) caption. 
To learn discriminative membership cues, we construct paired examples consisting of tracks and their caption-conditioned generations produced by shadow models, and use member and non-member sets to train a lightweight music auditor. The auditor learns alignment patterns characteristic of training membership and can be applied to the target model without access to its parameters or training data.


We evaluate our framework on three representative text-to-music systems (i.e., AudioLDM2~\cite{liu2024audioldm}, Stable Audio~\cite{evans2025stable}, and Mustango~\cite{melechovsky2024mustango}) under a leave-one-generator-out protocol using five complementary audio encoders. Under the best configuration, our method achieves up to 98.6\% accuracy, with false-positive and false-negative rates of 1.9\% and 1.0\%, respectively. We further observe strong cross-generator transferability and high sample efficiency, exceeding 98\% accuracy with fewer than 100 training pairs in certain settings.

\vspace{-1mm}
\section{Related Work}
\vspace{-1mm}
\subsection{Music Generative Models}
Early research in music generation primarily focused on symbolic representations such as MIDI, with models like MusicVAE~\cite{roberts2018hierarchical} and MuseGAN~\cite{dong2018musegan} coordinating pitch, velocity, and duration to produce multi-track compositions. More recently, the field has shifted toward text-conditioned models that synthesize audio directly from natural language prompts. Current academic approaches largely fall into three paradigms: diffusion-based models (e.g., AudioLDM2~\cite{liu2024audioldm}, Riffusion~\cite{Forsgren_Martiros_2022}, Stable Audio Open~\cite{evans2025stable}, Mustango~\cite{melechovsky2024mustango}) that iteratively denoise acoustic representations; transformer-based models (e.g., MusicGen~\cite{copet2023simple}, Lyria Realtime~\cite{lyriateam2025livemusicmodels}) that perform sequence modeling over discrete audio tokens; and flow-matching models (e.g., JASCO~\cite{tal2024joint}) that bridge noise and music distributions while incorporating symbolic constraints.
Beyond academia, proprietary systems such as Suno~\cite{sunoai2023} and Udio~\cite{udio2024} further demonstrate the ability to generate high-fidelity, multi-minute compositions with coherent long-term structure, though their internal architectures remain undisclosed.

\vspace{-2mm}
\subsection{Membership Inference}
Membership Inference Attacks (MIA) are typically categorized by the auditor's access level. In white-box settings, the auditor can inspect model parameters and internal states, whereas in black-box settings, interaction is limited to queries and returned outputs~\cite{sablayrolles2019white}.
In the image domain, membership inference and provenance auditing for generative models have been widely studied under realistic black-box access. Wu \textit{et al.}~\cite{wu2022membership} show that member samples induce generations that are closer in embedding space and more semantically aligned with the conditioning caption than non-members. Huang \textit{et al.}~\cite{huang2024general,huang2025instance} propose statistically grounded auditing frameworks that convert membership scores into formal decisions with false-discovery control, extending to instance-level auditing. Bohacek \textit{et al.}~\cite{bohacek2025genai} observe that member generations remain consistently closer to the source image across varying diffusion strengths. Zhu \textit{et al.}~\cite{zhu2025auditing} exploit semantic consistency among prompts, reference images, and generated outputs, reporting stronger alignment for member pairs.

Turning to generative audio and music, recent work has begun examining privacy risks in generative music models. Chow \textit{et al.}~\cite{chow2025assessing} audit MuseGAN under both white-box and black-box MIAs; in the black-box setting, their Monte Carlo output-similarity attack (e.g., Euclidean-distance based) achieves near-chance single-sample accuracy. Proboszcz \textit{et al.}~\cite{proboszcz2025membership} study MIAs on open-source audio generators using standard train/test splits as member and non-member sets. They report near-chance per-sample performance on large-scale models such as AudioLDM2~\cite{liu2024audioldm}, with improved results on smaller models like TANGO~\cite{deepanway2023text}, and enhance detection power via dataset-level statistical aggregation.
In parallel, Liu \textit{et al.}~\cite{liu2026ts} investigate symbolic music generators using teacher-forcing likelihoods, which require likelihood access and are impractical when only generated samples are observable. Liu \textit{et al.}~\cite{liu2026mi} propose a developer-side white-box attack on diffusion-based music models, scoring membership via intermediate-state perturbations needed to induce perceptual degradation.
Collectively, these studies highlight both the emerging importance and the difficulty of membership inference in generative music. Existing black-box approaches rely on limited scalar similarity statistics or dataset-level aggregation, while white-box methods assume internal access unavailable in deployed systems. 
\vspace{-2mm}
\section{System Design}
\vspace{-2mm}
\subsection{Threat Model}
\label{threatmodel}
\begin{figure*}[t]
    \centering
    \includegraphics[width=0.95\textwidth]{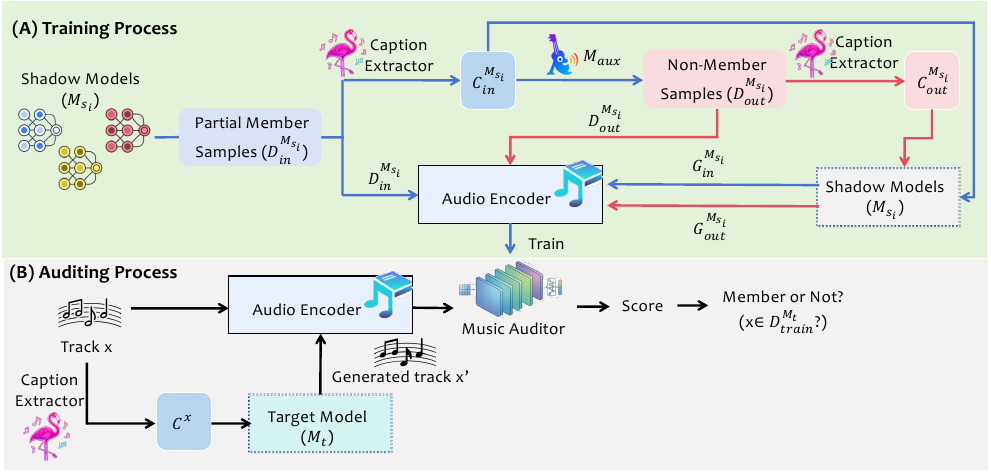}
    \caption{Overview of the proposed black-box membership inference auditing framework for generative music models.} 
    \vspace{-2mm}
    \label{fig:System Architecture}
    \vspace{-2mm}
\end{figure*}

\noindent\textbf{Target Model Owner.}
The model owner (e.g., an AI company or platform provider) trains a generative music model $\mathcal{M}_t$ (\textit{target model}) on a proprietary dataset $\mathcal{D}_{\mathrm{train}}^{\mathcal{M}_t}$ consisting of paired music-caption samples. The owner has full access to the training corpus, model parameters, and computational resources required for large-scale development and deployment. The trained system is exposed via an API that accepts text prompts and returns generated music.
However, the owner does not disclose the composition of $\mathcal{D}_{\mathrm{train}}^{\mathcal{M}_t}$, internal model parameters, or any mechanism for externally verifying whether a specific audio sample was included during training.

\noindent\textbf{Auditor Goal.}
An auditor (e.g., a musician, researcher, or regulatory authority) seeks to determine whether a candidate music sample $x$ was included in $\mathcal{D}_{\mathrm{train}}^{\mathcal{M}_t}$. Formally, the goal is to infer the membership indicator $m(x) \in \{0,1\}$, where $m=1$ denotes inclusion in the training set and $m=0$ otherwise. The auditing process is strictly post hoc and does not influence the training or operation of the target model.



\noindent\textbf{Auditor Capabilities.}
We assume the auditor has black-box query access to the target model $\mathcal{M}_t$, allowing them to submit a prompt and obtain the corresponding generated audio output. The auditor has no access to the target model's training set $\mathcal{D}_{\mathrm{train}}^{\mathcal{M}_t}$, nor to any internal information such as parameters, gradients, likelihood scores, or hidden activations. 
To learn transferable membership signals, the auditor may access several other text-to-music generative models $\{\mathcal{M}_{s_1},\ldots,\mathcal{M}_{s_n}\}$, referred to as \textit{shadow models}, for which verified subsets of training data are available. 
For each shadow model $\mathcal{M}_{s_i}$, the auditor uses known member samples $\mathcal{D}_{\mathrm{in}}^{\mathcal{M}_{s_i}} \subseteq \mathcal{D}_{\mathrm{train}}^{\mathcal{M}_{s_i}}$ to construct member and non-member auditing pairs and train a membership classifier (\textit{music auditor}).
The auditor also has access to computational resources as well as pre-trained audio encoders, music caption extraction models, and additional text-to-music generators ($\mathcal{M}_{\text{aux}}$, distinct from the target and shadow models). These tools are used to extract semantic and acoustic representations, obtain captions for candidate audio when needed, and generate semantically aligned non-member samples.
In addition,
we consider a strict black-box auditing setting in which the model owner does not deploy adaptive membership defenses. The auditing process is purely post-hoc and does not modify the target model or its training data.
\vspace{-2mm}
\subsection{System Overview}
\label{subsec:overview}
Figure~\ref{fig:System Architecture} illustrates our end-to-end auditing pipeline, which consists of a training process on shadow models and an auditing process on the target model.

\noindent\textbf{Training Process.}
Our key hypothesis is that training membership induces stronger semantic and structural alignment between a track and the model's generation conditioned on its caption.
To learn discriminative membership signals from shadow models, we assume access to $n$ shadow models $\{\mathcal{M}_{s_i}\}_{i=1}^{n}$ and partial verified member samples $\mathcal{D}^{\mathcal{M}_{s_i}}_{\text{in}}$ for each $\mathcal{M}_{s_i}$. 
Since models are prompt-conditioned, we first extract captions
$\mathcal{C}^{\mathcal{M}_{s_i}}_{\text{in}}$ from $\mathcal{D}^{\mathcal{M}_{s_i}}_{\text{in}}$ using a caption extractor (e.g., a large audio-language model). 
To obtain non-member audio that is comparable in semantics to avoid trivial separability, we condition an auxiliary text-to-music generative model $\mathcal{M}_{\text{aux}}$ on $\mathcal{C}^{\mathcal{M}_{s_i}}_{\text{in}}$ to construct semantically matched non-member samples $\mathcal{D}^{\mathcal{M}_{s_i}}_{\text{out}}$. We then extract prompts $\mathcal{C}^{\mathcal{M}_{s_i}}_{\text{out}}$ from $\mathcal{D}^{\mathcal{M}_{s_i}}_{\text{out}}$ using the caption extractor and query the shadow model to obtain corresponding generations
$\mathcal{G}^{\mathcal{M}_{s_i}}_{\text{in}} = \mathcal{M}_{s_i}(\mathcal{C}^{\mathcal{M}_{s_i}}_{\text{in}})$ and
$\mathcal{G}^{\mathcal{M}_{s_i}}_{\text{out}} = \mathcal{M}_{s_i}(\mathcal{C}^{\mathcal{M}_{s_i}}_{\text{out}})$.
This construction yields four aligned groups per shadow model, namely
$\mathcal{D}^{\mathcal{M}_{s_i}}_{\text{in}}$,
$\mathcal{D}^{\mathcal{M}_{s_i}}_{\text{out},i}$,
$\mathcal{G}^{\mathcal{M}_{s_i}}_{\text{in}}$, and
$\mathcal{G}^{\mathcal{M}_{s_i}}_{\text{out}}$, enabling controlled comparison between in-training and out-of-training behavior. These paired examples form the core supervision signal for learning discriminative alignment patterns that generalize across models.

To operationalize this, each audio sample is embedded using an encoder $f(\cdot)$, 
which abstracts away low-level waveform variability and maps audio into a 
semantically structured representation space. For each shadow model $\mathcal{M}_{s_i}$, we construct \emph{member pairs} 
$(d_{\text{in}}, g_{\text{in}})$, where 
$d_{\text{in}} \in \mathcal{D}^{\mathcal{M}_{s_i}}_{\text{in}}$ and 
$g_{\text{in}} \in \mathcal{G}^{\mathcal{M}_{s_i}}_{\text{in}}$, 
and \emph{non-member pairs} 
$(d_{\text{out}}, g_{\text{out}})$, where 
$d_{\text{out}} \in \mathcal{D}^{\mathcal{M}_{s_i}}_{\text{out}}$ and 
$g_{\text{out}} \in \mathcal{G}^{\mathcal{M}_{s_i}}_{\text{out}}$. Each pair is mapped into a fixed-dimensional representation (e.g., via concatenation in embedding space) and used to train a lightweight 
music auditor that captures alignment differences indicative of training membership.


\noindent\textbf{Auditing Process.} 
To audit a black-box target model $\mathcal{M}_t$ on a candidate track $x$, we extract its caption $C^x$, query the model to obtain $x' = \mathcal{M}_t(C^x)$, compute embeddings $f(x)$ and $f(x')$, and feed the pair into the trained auditor to obtain a membership score. A high score indicates that the alignment between $x$ and its generation resembles patterns learned from shadow-model members, suggesting likely inclusion in training.

\vspace{-2mm}
\subsection{Semantic Evidence of Membership}
To examine whether training membership induces stronger alignment between a track and its prompt-conditioned generation, we conduct a preliminary model-based assessment using Gemini~2.5~Flash. We use Stable Audio Open~\cite{evans2025stable} as the generative model and randomly sample 100 member tracks from the documented training set. Following Section~\ref{subsec:overview}, we construct an evaluation set consisting of 100 member pairs and 100 semantically matched non-member pairs. The scoring prompt is provided in Table ~\ref{tab: score prompt of gemini2.5}.
For each pair, Gemini outputs a similarity score on a 0--10 scale with a brief rationale. Member pairs achieve a higher average score (mean $=7.44$) than non-member pairs (mean $=6.94$), indicating stronger perceived correspondence in the member setting. 

While member pairs exhibit higher similarity on average, the gap remains moderate because prompt-level conditioning already enforces strong global semantic alignment in both groups; membership primarily contributes finer-grained structural consistency that is only partially captured by a single scalar similarity score. This analysis is intended as a preliminary study rather than a definitive mechanistic conclusion. Its main purpose is to examine what kind of semantic-alignment patterns emerge when using recent LLM-based captioning, rather than to demonstrate a statistically significant gap between member and non-member cases.


\begin{table}[t]
\centering
\caption{Semantic evaluation prompt for Gemini 2.5.}
\label{tab: score prompt of gemini2.5}

\vspace{-1mm}
\resizebox{0.9\linewidth}{!}{%
\begin{minipage}{\linewidth}
\begin{tcolorbox}[
  enhanced,
  width=\linewidth,
  boxsep=1pt,
  top=1pt,
  bottom=1pt,
  left=2pt,
  right=2pt,
  arc=1mm,
  colback=white,
  colframe=darkblue,
  colbacktitle=darkblue,
  coltitle=white,
  title={\small \textbf{Gemini 2.5 Flash Prompt}}
]
System Role: You are an expert Audio Forensics Analyst specializing in Machine Learning evaluation and music plagiarism detection. Your goal is to determine if "File B" (the AI generation) exhibits signs of training data leakage or overfitting based on "File A" (the potential training data). \\

Task: Compare the two provided audio files. Disregard file duration, format, path, bitrate, loudness differences. Focus exclusively on compositional and timbral identity.\\

Analyze the following specific elements:

Melodic Contour: Are the main melodic motifs or lead lines identical or near-identical?

Harmonic Structure: Is the chord progression and specific voicing the same?

Timbral Specificity: Does the AI attempt to replicate specific instrument sounds, synth patches, or vocal textures found in the original?

Structural Quirks: Are there unique rhythmic patterns, drum fills, or production artifacts (like vinyl crackle or specific glitches) from the original that appear in the generation? \\

Output Requirements:

Provide a brief analysis of the elements above.

Similarity Score (0-10):

0-3: Different songs, same genre (Normal model behavior).

4-7: Strong stylistic copying, but different composition (Style transfer).

8-10: Clear evidence of overfitting/memorization (The AI is reproducing the training data).

Return ONLY the following JSON (no extra text):\\
  "score": number 0-10,\\
  "key\_factor": "brief reason"
\end{tcolorbox}
\vspace{-6mm}
\end{minipage}%
}
\end{table}

Table~\ref{tab:gemini_exmaples} provides examples of semantic comparison of member and non-member pairs. We observed that on the member pair, the audio-language model clearly identifies their similarity and gives a score of 10, which confirms direct reproduction of melodic and timbral elements. Conversely, it gives a score of 0 to the non-member pair, correctly recognizing the lack of any resemblance. 
\begin{table}[t]
\centering
\vspace{2mm}
\caption{Semantic comparison examples.}
\label{tab:gemini_exmaples}
\resizebox{0.9\linewidth}{!}{%
\begin{minipage}{\linewidth}
\begin{tcolorbox}[
  enhanced,
  width=\linewidth,
  boxsep=1pt,
  top=1pt,
  bottom=1pt,
  left=2pt,
  right=2pt,
  arc=1mm,
  colback=white,
  colframe=darkblue,
  colbacktitle=darkblue,
  coltitle=white,
  title={\small \textbf{Example output (Member; score = 10)}}
]
\textbf{key\_factor:} "Direct reproduction of multiple distinct melodic phrases, specific timbral elements (synth patches, percussive sounds), and structural progressions from the source material. The AI generation is an almost identical, albeit condensed, version of the training data."
\end{tcolorbox}
\begin{tcolorbox}[
  enhanced,
  width=\linewidth,
  boxsep=1pt,
  top=1pt,
  bottom=1pt,
  left=2pt,
  right=2pt,
  arc=1mm,
  colback=white,
  colframe=darkblue,
  colbacktitle=darkblue,
  coltitle=white,
  title={\small \textbf{Example output (Non-member; score = 0)}}
]
\textbf{key\_factor:} "File B is a completely different musical composition with distinct melodic, harmonic, and timbral characteristics, showing no resemblance to File A."
\end{tcolorbox}
\vspace{-6mm}
\end{minipage}%
}
\end{table}

\vspace{-2mm}
\section{Methodology}
\vspace{-2mm}

\subsection{Data Generation} 
\label{sec: data_generation}

\begin{table}[t]
\centering
\caption{Example captions for Music Flamingo.}
\label{tab:music_flamingo_prompts}
\vspace{-2mm}
\resizebox{0.9\linewidth}{!}{%
\begin{minipage}{\linewidth}
\begin{tcolorbox}[
  enhanced,
  width=\linewidth,
  boxsep=1pt,
  top=1pt,
  bottom=1pt,
  left=2pt,
  right=2pt,
  arc=1mm,
  colback=white,
  colframe=darkblue,
  colbacktitle=darkblue,
  coltitle=white,
  title={\small \textbf{Example Caption 1}}
]
This is a blues-influenced rock track in Eb major with a moderate tempo of 142.86 BPM. It features a steady 4/4 rhythm with a chord progression alternating between G\# major and D\# major, creating a soulful and laid-back atmosphere. The instrumentation likely includes electric guitar, bass, and drums, delivering a classic blues-rock vibe.
\end{tcolorbox}

\vspace{-2mm}

\begin{tcolorbox}[
  enhanced,
  width=\linewidth,
  boxsep=1pt,
  top=1pt,
  bottom=1pt,
  left=2pt,
  right=2pt,
  arc=1mm,
  colback=white,
  colframe=darkblue,
  colbacktitle=darkblue,
  coltitle=white,
  title={\small \textbf{Example Caption 2}}
]
This indie rock track features energetic electric guitars with a driving rhythm section, including punchy drums and a steady bassline. The vocals are raw and emotive, delivering introspective lyrics with a gritty edge. The overall mood is intense and rebellious, typical of mid-2000s alternative rock.
\end{tcolorbox}
\end{minipage}%
}
\end{table}

To train and evaluate the music auditor, we collect four types of audio samples: member and non-member tracks, along with their corresponding generations produced by either the target model or shadow models.
To ensure realistic evaluation and verifiable membership ground truth, the member set must satisfy two strict criteria: (1) tracks must be explicitly documented as part of the model's training corpus, and (2) complete metadata (e.g., track names and sources) must be available to guarantee reliable data provenance. Following these criteria, we select three pre-trained text-to-music generative models, i.e., AudioLDM2~\cite{liu2024audioldm}, Stable Audio Open~\cite{evans2025stable}, and Mustango~\cite{melechovsky2024mustango}, which have publicly documented training datasets.

For AudioLDM2 and Stable Audio Open, the Free Music Archive (FMA)~\cite{defferrard2016fma} constitutes a documented subset of their broader training corpora. FMA is a large-scale dataset containing over 106,000 tracks with rich metadata, enabling reliable provenance verification. AudioLDM2 uses FMA as part of the data for training its VAE and jointly fine-tuning its GPT-2~\cite{radford2019language} and Transformer-UNet, while Stable Audio Open leverages FMA to train its autoencoder and diffusion transformer. 
Mustango is trained on MusicBench~\cite{melechovsky2024mustango} (52,710 samples), which is augmented from MusicCaps~\cite{agostinelli2023musiclm} and serves as the training source for its modified UNet (MuNet).
For member selection, we randomly sample 1,000 2-3 minute tracks from FMA for AudioLDM2 and Stable Audio Open, manually verifying that no overlap exists between their subsets. For Mustango, we retrieve 1,000 ten-second tracks from MusicBench.

To construct the non-member sets, we ensure that each member–non-member pair is semantically aligned in content, so that any observed differences between in-training and out-of-training samples cannot be attributed to semantic variation (e.g., genre, instrumentation). This design isolates training exposure as the primary distinguishing factor.
Concretely, we first extract detailed descriptive captions from each member track using Music Flamingo~\cite{ghosh2025music}. Example captions can be found in Table~\ref{tab:music_flamingo_prompts}. These captions summarize high-level musical attributes and compositional characteristics. We then provide the captions to Jam~\cite{liu2025jam}\footnote{We select Jam because it supports long-duration synthesis while maintaining strong perceptual fidelity, fewer audible artifacts, and faster generation compared to alternatives such as YUE~\cite{yuan2025yue}.}, a high-fidelity text-to-music generator, to generate corresponding non-member tracks conditioned on the same semantic description.

After constructing the non-member set, we apply the same caption extraction pipeline to each non-member track, as illustrated in Section~\ref{subsec:overview}, to obtain its corresponding textual description. Using the captions derived from both the member and non-member sets, we query the music generation models to generate conditioned outputs. This procedure yields two types of paired samples: (1) member tracks and their corresponding generations, and (2) non-member tracks and their corresponding generations. These paired samples are then used to train and evaluate the music auditor. Due to decoding/generation errors, 
the final dataset sizes vary slightly across models. In total, we obtain 995 member-generation pairs and 995 non-member-generation pairs for AudioLDM2, 1,000 pairs of each type for Stable Audio Open, and 989 member-generation pairs and 991 non-member-generation pairs for Mustango.

\vspace{-2mm}
\subsection{Audio Encoders}
\label{sec: audio encoders}
\begin{table}[t]
  \centering
  \caption{Summary of audio encoders and aggregation.}
  \label{tab:feature_summary}
  \footnotesize 
  \rowcolors{2}{white}{lightblue}
  \renewcommand{\arraystretch}{1.4}
  \begin{tabular}{>{\centering\arraybackslash}m{1.6cm}
                  >{\centering\arraybackslash}m{2.0cm}
                  >{\centering\arraybackslash}m{1.7cm}
                  >{\centering\arraybackslash}m{1.8cm}}
    \rowcolor{darkblue}
    \color{white}\textbf{Model} &
    \color{white}\textbf{View} &
    \color{white}\textbf{Raw Output} &
    \color{white}\textbf{Aggregation} \\

    MERT~\cite{yizhi2023mert} & Music semantics & $\mathbb{R}^{L \times T \times D}$ & Mean over $T$ \\

    Music2Vec~\cite{li2022map} & Structure/tonality & $\mathbb{R}^{L \times T \times D}$ & Mean over $T$ \\

    DAC~\cite{kumar2023high} & Acoustic artifacts & $\{0..V\}^{N_q \times T}$ & Histogram over $V$ per $N_q$ \\

    Fx-Encoder++~\cite{yeh2025fx} & Production effects & $\mathbb{R}^{128}$ & Identity \\

    CLAP~\cite{wu2023large} & Audio–text semantics & $\mathbb{R}^{512}$ & Identity \\

    \bottomrule
  \end{tabular}
  \vspace{-2mm}
\end{table}
To quantify membership-related discrepancies between member, non-member, and generated audio samples, we employ multiple audio encoders, as shown in Table~\ref{tab:feature_summary}, and systematically compare the extent to which their representations retain training data signals.

\textbf{MERT}~\cite{yizhi2023mert} is a self-supervised Transformer trained with acoustic and musical teachers (e.g., CQT) to capture high-level musical semantics such as pitch and harmony. Using the 330MB checkpoint, we extract hidden states from all layers and apply temporal mean pooling within each layer, retaining the layer dimension to obtain fixed-size (Layer, Embedding) representations for membership analysis.

\textbf{Music2Vec}~\cite{li2022map} adopts a Data2Vec-style training objective with a CNN front-end and Transformer encoder to learn structured tonal representations. We extract frame-level hidden states and perform temporal mean pooling per layer, yielding duration-invariant (Layer, Embedding) features.

\textbf{DAC}~\cite{kumar2023high} is used to probe low-level acoustic memorization. We extract discrete codebook index sequences from its residual vector quantization layers and compute normalized histograms for each codebook. The concatenated histograms form a global acoustic fingerprint reflecting distributional acoustic patterns.

\textbf{Fx-Encoder++}~\cite{yeh2025fx} is adopted to assess production-style memorization. It disentangles audio effects from musical content and produces a global embedding representing mixing and production characteristics.

\textbf{CLAP}~\cite{wu2023large} (HTSAT-base, non-fusion) provides a 512-dimensional global embedding from the full 48 kHz mono track, enabling evaluation of membership signals within high-level semantic representations learned from large-scale audio–text supervision.
\vspace{-1mm}
\subsection{Music Auditor}
\label{sec: music auditor}
\vspace{-1mm}
After extracting audio representations, we train a light-weight \textit{Music Auditor} to classify member vs.\ non-member from embedding-space evidence. Because different encoders output different representation formats and dimensionalities, we use two auditor backbones: a multi-layer perceptron (MLP) for globally pooled 1D embeddings and a convolutional neural network (CNN) for 2D embedding maps that preserve structured axes. For each encoder, we instantiate the auditor with an input layer matched to that encoder's embedding size.

 Given a member/non-member--generated pair of 1D embeddings $(\big(f(x), f(x')\big)\in\mathbb{R}^{d}$, we form the pair feature by concatenation $[f(x)\, \|\, f(x')]\in\mathbb{R}^{2d}$ and feed it into an MLP. The MLP network has three hidden layers of sizes 256, 128, and 64 with ReLU activations, followed by a linear output layer that produces a single logit. When an encoder outputs a 2D embedding map $\big(f(x), f(x')\big)\in\mathbb{R}^{L\times D}$ (e.g., preserving a layer/temporal axis), we construct paired inputs by stacking the two maps along the channel dimension, yielding a tensor of shape $(2,L,D)$. The CNN network contains three $3\times3$ convolution layers with 32, 64, and 128 channels (padding 1), each followed by ReLU. We apply adaptive average pooling to obtain a fixed-length representation and use a final fully connected layer to output a single logit. We train the two auditors using BCEWithLogitsLoss and optimize with AdamW, with early stopping based on validation loss.

\vspace{-2mm}
\section{Evaluation}
\begin{figure*}[!t]
    \centering
    \includegraphics[
        width=0.95\textwidth,
        trim={0 2pt 0 0},
        clip
    ]{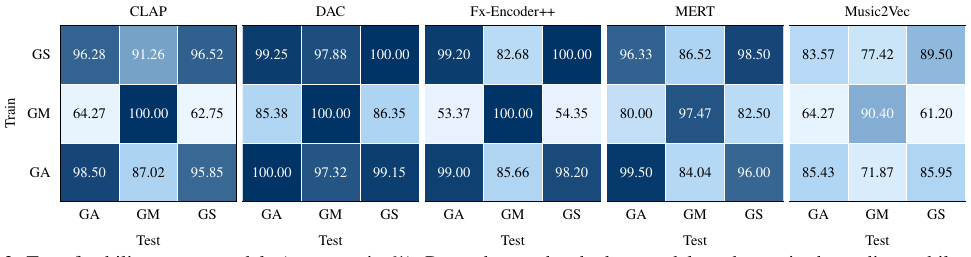}
    \caption{Transferability across models (accuracy in \%). Rows denote the shadow model used to train the auditor, while columns denote the target model used for evaluation. GA = AudioLDM2, GM = Mustango, and GS = Stable Audio Open.}
    \label{fig:transferability}
    \vspace{-2mm}
\end{figure*}

\subsection{Experimental Settings}
\noindent\textbf{Text-to-music Models.} As discussed in Section~\ref{sec: data_generation}, we select three text-to-music models for our evaluation: AudioLDM2~\cite{liu2024audioldm}, Stable Audio Open~\cite{evans2025stable}, and Mustango~\cite{melechovsky2024mustango}.
To ensure authenticity and reproducibility, we use the publicly available checkpoints for all three models and retain their default hyperparameter settings during generation. We set the output duration to 30 seconds for AudioLDM2 and Stable Audio Open, and to 10 seconds for Mustango. 
For non-member generation, we employ the auxiliary model Jam~\cite{liu2025jam}
to generate instrumental tracks while keeping its default hyperparameters unchanged. To ensure fair comparison, we match the non-member duration to the corresponding member duration for each target generator. Specifically, the non-member tracks are standardized to 150 seconds for Stable Audio Open and AudioLDM2, and to 10 seconds for Mustango.


\noindent\textbf{Audio Pre-processing.}
Before feature extraction, we standardize all audio by transcoding it to MP3 at a fixed bitrate (256 kbps), then downmixing to mono and resampling to the target sampling rate required by each encoder.

Unless mentioned otherwise, we evaluate our pipeline under a three-way leave-one-generator-out protocol over AudioLDM2, Mustango, and Stable Audio Open. In each run, one generator is held out as the \emph{target model} for testing, while the remaining two act as \emph{shadow models} for training the music auditor. Concretely, for each \emph{shadow model}, we split its member/non-member pairs into train/validation sets with an 80\%/20\% ratio; the two \emph{shadow models} splits are then pooled to train a single auditor.
During evaluation, we test on all available samples from the held-out \emph{target model}, without further subsampling.

\vspace{-2mm}
\subsection{Experimental Results}
\subsubsection{Main Results}
\begin{table}[t]
\centering
\scriptsize
\setlength{\tabcolsep}{2.5pt}
\renewcommand{\arraystretch}{1.25} 
\setlength{\aboverulesep}{0pt}
\setlength{\belowrulesep}{0pt}
\rowcolors{3}{lightblue}{white} 

\begin{tabular}{l ccc ccc ccc c}
\toprule

\rowcolor{darkblue}
\color{white}\textbf{Encoder} &
\multicolumn{3}{c}{\color{white}\textbf{AudioLDM2}} &
\multicolumn{3}{c}{\color{white}\textbf{Mustango}} &
\multicolumn{3}{c}{\color{white}\textbf{StableAudioOpen}} &
\color{white}\textbf{Avg} \\

\rowcolor{white} 
& \textbf{Acc} & \textbf{FPR} & \textbf{FNR}
& \textbf{Acc} & \textbf{FPR} & \textbf{FNR}
& \textbf{Acc} & \textbf{FPR} & \textbf{FNR}
& \textbf{Acc} \\
\midrule

MERT          & 95.6 & 1.9  & 6.9  & 84.0 & 11.1 & 20.8 & 95.3 & 2.1  & 7.2  & 91.7 \\
Music2Vec     & 83.9 & 24.5 & 7.7  & 75.2 & 20.2 & 29.3 & 75.6 & 30.0 & 18.7 & 78.3 \\
DAC           & \textbf{97.5} & \textbf{0.0}  & \textbf{5.0}  & \textbf{97.1} & \textbf{5.5}  & \textbf{0.4}  & \textbf{98.6} & 1.9  & \textbf{1.0}  & \textbf{97.7} \\
Fx-Encoder++  & 69.4 & \textbf{0.0}  & 61.2 & 86.6 & 24.2 & 2.6  & 94.5 & \textbf{0.9}  & 10.0 & 83.5 \\
CLAP          & 89.6 & 3.0  & 17.8 & 90.0 & 13.4 & 6.5  & 95.5 & 2.5  & 6.5  & 91.7 \\

\midrule

\rowcolor{white} 
\textbf{Average}
& 87.2 & 5.9  & 19.7
& 86.6 & 14.9 & 11.9
& \textbf{91.9} & 7.5 & 8.7
& {} \\ 

\bottomrule
\end{tabular}
\caption{Accuracy evaluation (\%) of five music encoders on each target music generator. For each column, the best-performing encoder is highlighted in bold.}
\label{tab:three_tests_singlecol}
\vspace{-3mm}
\end{table}

We first evaluate whether black-box membership auditing is feasible using embedding-based signals, and which audio encoder yields the most discriminative representations for separating members from non-members. We extract embeddings from five encoders, train auditors on the pooled shadow models, and report Accuracy (Acc), False Positive Rate (FPR), and False Negative Rate (FNR) on the held-out target model (Table~\ref{tab:three_tests_singlecol}). Among all encoders, DAC is the most effective and stable, achieving the best average Acc of 97.7\% and remaining strong across all targets, with near-saturated performance on Stable Audio Open (98.6\% Acc, FPR 1.9\%, FNR 1.0\%) and Mustango (97.1\% Acc, FNR 0.4\%), and similarly high accuracy on AudioLDM2 (97.5\% Acc; FPR 0.0\%, FNR 5.0\%). In contrast, Music2Vec is consistently weaker (78.3\% average Acc) and exhibits notably higher false-positive rates across all targets (20--30\% FPR). Fx-Encoder++ shows pronounced cross-target instability, collapsing on AudioLDM2 (69.4\% Acc) due to a very high false-negative rate (61.2\%) despite strong results on Stable Audio Open (94.5\% Acc). Finally, averaging across encoders, Stable Audio Open yields the highest target-level Acc (91.9\%), suggesting that membership signals are most detectable for this model.

\subsubsection{Transferability Analysis}
We also evaluate cross-model transferability by training on a single model and testing on the other two. Concretely, for a chosen source model, we split its pairs into 70\%/20\%/10\% for train/validation/test, where the 10\% split forms an \emph{in-domain} test set. We then evaluate the same trained classifier on all pairs from each of the remaining models, yielding two \emph{cross-model} test sets. The results from Figure~\ref{fig:transferability} show that in-domain separation is generally strong for CLAP, DAC, Fx-Encoder++, and MERT (diagonal Acc mostly in the mid-to-high 90\%s, often $\approx$100\%), while Music2Vec is consistently weaker (diagonal Acc in the mid-80\%s to low-90\%s). Cross-model generalization is most stable between AudioLDM2 and Stable Audio Open: CLAP/DAC/Fx-Encoder++/MERT remain high in both directions (typically $\approx$96-100\%), whereas Music2Vec drops to the mid/high-80\%s.

Mustango primarily introduces an asymmetric transfer gap: when the auditor is \emph{trained} on Mustango, several encoders fail to transfer to the long-form generators (e.g., CLAP and Fx-Encoder++ drop to $\sim$53-64\% when testing on AudioLDM2/Stable Audio Open), with Music2Vec showing a similar but milder degradation ($\sim$61-64\%). In contrast, DAC is comparatively robust under the same shift (still $\sim$85--86\%), and MERT degrades more moderately ($\sim$80-83\%). When Mustango is used only as the \emph{test} generator, the degradation is less severe for most encoders (DAC remains $\sim$97-98\%, CLAP $\sim$87-91\%, and Fx-Encoder++/MERT in the mid-80\%s), while Music2Vec is the most affected (low-to-mid 70\%s). This pattern is consistent with a distribution shift driven by Mustango's short segments ($\sim$10\,s) versus the long-form pairs used for AudioLDM2 and Stable Audio Open ($\sim$150\,s), which can make auditors trained on short clips rely on cues that do not generalize to long-form audio.

\subsubsection{Impact of Training Size}
\begin{figure}[t]
    \centering
    \includegraphics[width=0.9\linewidth]{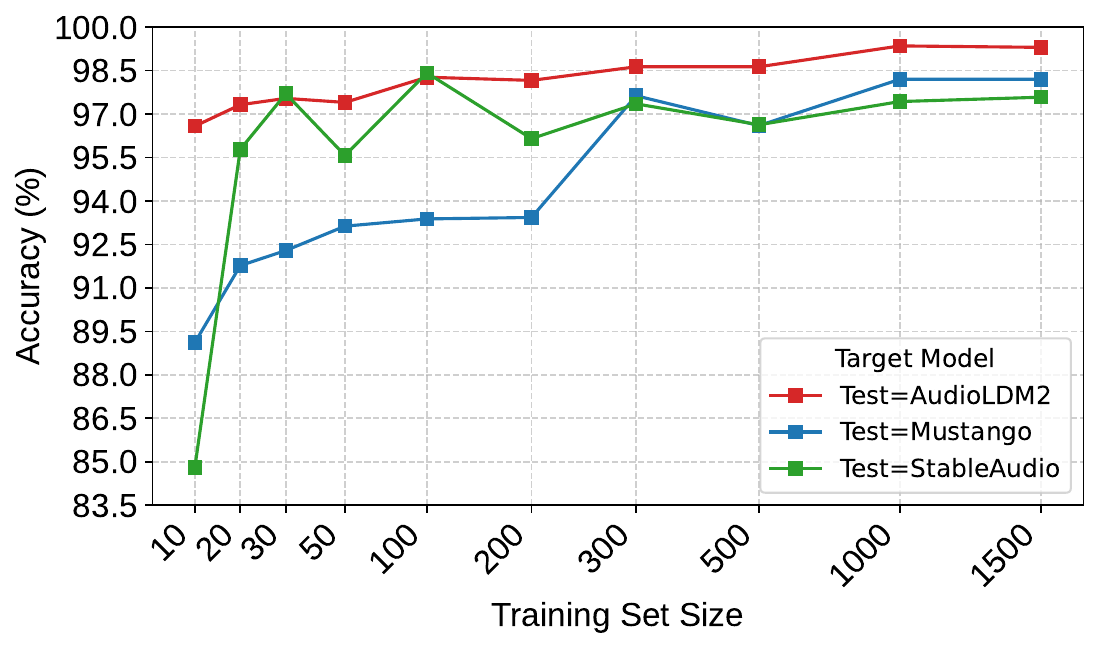}
    \caption{Training-set ablation using DAC across leave-one-generator-out settings.}
    \label{fig:ablation}
    \vspace{-2mm}
\end{figure}

We further conduct a training-set ablation study to quantify how many pairs are required to learn an effective auditor. Anchoring on the best-performing default configuration (DAC), we evaluate each leave-one-generator-out setting (holding out AudioLDM2, Mustango, or Stable Audio Open). For each training size, we randomly subsample pairs from the pooled shadow-model training data, repeat this process $k{=}3$ times, and report the mean performance (Figure~\ref{fig:ablation}). The validation split is fixed, and early stopping is applied based on validation loss.
We observe a rapid sample-efficiency regime followed by saturation. When AudioLDM2 is held out, accuracy reaches 96.6\% at $n{=}10$, exceeds 98\% by $n{=}100$, and peaks at 99.3\%. Stable Audio Open shows a similar pattern, rising from 84.8\% at $n{=}10$ to above 98\% by $n{=}100$ and stabilizing in the high-90\% range. In contrast, Mustango improves more gradually and requires roughly $n{\approx}1000$ pairs to surpass 98\%, indicating a stronger cross-generator shift.
\vspace{-2mm}
\section{Conclusion \& Discussion}
In this work, we show that training membership is associated with a measurable semantic and structural consistency between a candidate track and its caption-conditioned generation from the target model. Our current experiments are conducted under a partially informed shadow-model setting, where we assume access to a subset of member samples together with their ground-truth captions and metadata. Under this setting, the proposed auditor is designed to learn membership-associated alignment signals rather than robustness to metadata absence. By leveraging paired shadow-model examples and learning discriminative alignment patterns in feature space, our auditor effectively distinguishes members from non-members without access to target model parameters or training metadata. Experiments across multiple generative models demonstrate strong accuracy and low error rates, suggesting that training-data auditing is practical and effective in realistic black-box deployment settings.


Despite these promising results, several limitations remain and point to important future directions. Our current evaluation is limited in scale and diversity ($\sim$1k members per shadow generator) and depends on a specific caption-extraction and auxiliary-generation pipeline, which may introduce distribution bias. Moreover, during the training phase, the non-member case is modeled using a replacement hold-out sample rather than no sample, because the auditor requires pair-structured inputs for both classes. This is a practical approximation in the black-box setting, but it may cause part of the learned signal to depend on how non-member controls are constructed. Generator-specific constraints, such as short outputs or instrumental-only generations, further narrow variability and likely contribute to reduced out-of-distribution robustness. Future work will expand dataset heterogeneity, reduce reliance on fixed auxiliary pipelines, and develop auditing procedures with principled statistical guarantees and calibrated false-positive control under realistic deployment conditions.

\section*{\centering \normalsize ACKNOWLEDGMENT}
We would like to thank our anonymous reviewers and shepherd for their insightful feedback. This work is supported in part by NSF CNS-2114161, ECCS-2132106, CBET-2130643, and CNS-2403529.
\bibliographystyle{ieeetr}
\bibliography{reference}

\IEEEpeerreviewmaketitle

\end{document}